# Exa-PSD: A new Persian sentiment analysis dataset on Twitter

**Seyed Himan Ghaderi**
himan_ghaderi@alumni.iust.ac.ir
**Saeed Sarbaziazad**
s.sarbaziazad@modares.ac.ir
**Mohammad Mehdi Jaziriyan**
m_jaziryan@modares.ac.ir
**Ahmad Akbari**
ahmadakbari@ut.ac.ir

**Abstract** Today, Social networks such as Twitter are the most widely used platforms for communication of people. Analyzing this data has useful information to recognize the opinion of people in tweets. Sentiment analysis plays a vital role in NLP, which identifies the opinion of the individuals about a specific topic. Natural language processing in Persian has many challenges despite the adventure of strong language models. The datasets available in Persian are generally in special topics such as products, foods, hotels, etc while users may use ironies, colloquial phrases in social media To overcome these challenges, there is a necessity for having a dataset of Persian sentiment analysis on Twitter. In this paper, we introduce the Exa sentiment analysis Persian dataset, which is collected from Persian tweets. This dataset contains 12,000 tweets, annotated by 5 native Persian taggers. The aforementioned data is labeled in 3 classes: positive, neutral and negative. We present the characteristics and statistics of this dataset and use the pre-trained Pars Bert and Roberta as the base model to evaluate this dataset. Our evaluation reached a 79.87 Macro F-score, which shows the model and data can be adequately valuable for a sentiment analysis system.

## 1 Introduction

The impact of online social networks, as well as their rapid growth, has revolutionized the patterns of human interaction and information distribution in recent years. Twitter is one of the most popular microblogs in the world enabling users to post their opinions, views, etc in the form of tweets having 280 or less characters. Twitter is a rapidly growing social media network in which users can "follow" other users to be notified when they have some activities.

Recognizing people's opinion from text is called sentiment analysis, that provides insight into people's attitudes of products, events, etc[1]. In social media, sentiment analysis can be used to determine whether the opinion of the writer of a text is negative, neutral or positive about an issue. It can be interesting for companies to know people's thoughts about their services [2]. The rapid and exponential data growth has made it very difficult for individuals, organizations, institutions and governments to make decisions based on knowledge and helping them to effectively benefit from the data. Manual sentiment analysis on such a large number of comments is almost impossible due to the rapid growth of contents and tweets. As a result, an automated system capable of sentiment analysis is necessary.

Text sentiment analysis is mostly done in three levels, called word or aspect, sentence and document levels [3].Aspect-level sentiment analysis involves determining the polarity of a specific target in the given context. In sentence and document levels, the overall polarity of a sentence and document are analyzed respectively. In other words, in document level the whole document is considered as an entity while in sentence level each sentence is considered as an entity and the sentiment polarity of the entity is analyzed.

Previously, most of the traditional methods in sentiment analysis were done using a lexicon-based method that determined the sentiment of a sentence based on the presence of some terms with positive or negative sentiment scores. In this technique each word's sentiment is ranked in a sentiment lexicon corpus, and the label of a sentence is the sum of the sentence lexicon scores. The idea behind this approach is that some lexicons mostly have intrinsic sentiment meanings[4]. Another approach for sentiment analysis is to use machine learning. Machine learning is categorized into two sub-approaches: supervised and unsupervised. The main characteristic of supervised learning is the availability of annotated training data; however, when there is no annotated corpus, unsupervised algorithms are applied.. Recently, deep neural networks have been used to do tasks in Natural Language Processing (NLP) such as sentiment analysis[5], name entity recognition[6], language modeling[7], etc. These models have revolutionized machine learning applications as well as prediction performance [8].

As aforementioned, the essence of supervised machine learning model creation is the existence of annotated training data. Most of the annotated NLP datasets are in rich-resource language with the goal of higher performance for industrial and organizational usage. Low-resource languages such as Persian, are not mostly well studied and suffer from lack of tools and resources. Although, having challenges in processing these low-resource languages, there is not a corpus having a high number of text (tweet) with a reliable tag. Most of the available Persian datasets are generally in special fields such as products, foods, hotels, etc. On the other hand, there are many compounds, informal and colloquial words, and idioms in Persian that makes the text processing more challenging in Persian. One of the ways for collecting data for low-resource languages is creating a corpus by annotating raw text.

In this study a new Persian sentiment analysis corpus of tweets is introduced. This corpus consists of some tweets being annotated manually in three sentiment classes. In the remainder of this paper the related works done in sentiment classification is summarized in section 2. The data gathering procedure is described in section 3 And finally experimental results and conclusions are discussed in sections 4, 5 respectively.

## 2 Background

One of the necessities for developing sentiment analysis with machine learning, is having an annotated corpus. There are some datasets for sentiment analysis but most of them are in English and few are available in low-resource languages such as Persian. In this section, some of the recent studies in sentiment analysis are discussed. Different approaches and available datasets are described in brief.

Bagheri et al. carried out two studies on Persian sentiment analysis [9][10]. They explained numerous aspects of the Persian language, such as verb suffixes, plural suffixes, and informal or colloquial phrases. Modified Mutual Information (MMI) was offered as a relevant feature that improved the Mutual Information (MI) metric. For sentiment analysis of Persian literature, Amiri et al[11] employed a sentiment lexicon-based technique From Persian online resources containing 7179 words, adjectives, expressions. They have studied issues affecting the sentiment polarity such as formal, informal, standard and obsolete terms.

Saraee in [12] uses n-gram, stemming and feature selection in the preprocessing phase. then a naive bayes classifier has been applied on data. The results show that the proposed method achieved 82.36% accuracy.

A new hybrid method for Persian sentiment analysis is proposed in [13] that uses integrated linguistic rules. In this method deep learning has been applied for classification. The results indicate that the proposed method outperformed other deep learning and also machine learning methods. in[14] a corpus having about 12000 Persian tweets in the field of the stock market keyword were manually annotated in three negative, neutral and positive labels. The result of the finetuned ParsBert on this data achieved 82% accuracy on test data.

Research [15] is one of the most current lexicon-based experiments in Persian that uses two lexicons to determine sentiment polarity. The documents were first parsed into Rhetorical Structure Trees for this investigation (RST). The scores were then summed using weighting rules.

In this domain, several algorithms and feature extraction approaches like lexicons, Part of Speech (POS), word embedding, character embedding, Term Frequency Inverse Document Frequency (tf-idf), etc have been used.

### 2.1 Available corpus and lexicons

Here, some Persian sentiment annotated corpuses are presented. These datasets are mostly comments of users in websites of digital products, hotels, food, etc. HesNegar

[16] is a Persian sentiment wordnet that is constructed using a lexi con network called FerdosNet and mapping from an English sentiment lexicon (SentiWordNet). This lexicon contains 100062 unique words and each word has a sentiment score.

In [17], Najafzadeh describes a self-training approach for dynamically extracting sentiment attributes. It is stated that employing colloquial n-gram as a heuristic to explore undiscovered and new issues in Persian language, can function instead of humans for extracting attributes. LexiPers [18] is another lexicon created based on the root terms from FarsNet. The words were manually annotated and then it was extended with a semi-supervised method. Some other datasets are listed in Table 1. The column indicates the name of the corpus, its level (word, sentence, document ), the size, domain and publisher of the corpus.

Table 1: some available corpuses on Persian sentiment Analysis

| Name | Level | Size | Domain | # Classes | Publisher |
|---|---|---|---|---|---|
| LexiPers | Word | 4261 | Public | 3 | [18] |
| SentiPers | Formal, informal or colloquial sentences | 26000 | Digital Product | 6 | [19] |
| PerSent | Lexicon, idiomatic words and expression | 1500 words &700idiomatic words and expression | Public | 3 | [20] |
| Persian lexicon with polarity label | word | 588 adjectives &4073 verbs &7325 nouns | Public | 3 | [21] |
| MirasOpinion | Document | 386 | Product | 3 | [22] |
| Hellokish | Document | 642 | Hotel | 2 | [23] |
| Hesnegar | word | 100062 | - | 2 | - |

## 3 Data collection and annotation

The new sentiment analysis dataset contains 12,000 Persian collected using the Twitter API prior to the recent policy changes implemented by Twitter. At the time of data collection, the API allowed for extensive access to public tweets, which facilitated the acquisition of the dataset analyzed in this research. However, it is important to note that Twitter's API policies have since been updated, and similar data collection may not be feasible under the current terms of service. For the annotation process, The Utag [1] annotation website has been used. The Utag website is developed by Exa [2] corporation for tagging and annotating data samples in natural language processing. This website is used for annotating single and multi-label in sentence level or token level annotation. The important feature of this website is that all annotations are done separately to prevent the tagger's mind bias. The collected tweets are annotated into three different classes based on the content of tweets the writer expressed in negative, neutral, or positive.

Table 2: dataset statistics details

| Final tag | Tag 1 | Tag 2 | Tag 3 | Tweet |
|---|---|---|---|---|
| Positive | Positive | Positive | Positive | Let me kiss you that you are so dear 😘😘😘❤️❤️<br>بیا ماچت کنم که تو اینقدر عزیزی😘😘😘❤️❤️ |
| Positive | Positive | Positive | Positive | I miss all of you, kind people of Twitter 😍<br>دلم برا همتون تنگ شده مهربونای توییتر 😍 |
| Negative | Negative | Negative | Negative | If I wasn't sick, I wouldn't be friends with you, hey<br>من اگه مریض نبودم با تو رفاقت نمیکردم هییییی |
| Negative | Negative | Negative | Negative | I wish I understood why I am so stressed and heartbroken today<br>ای کاش میفهمیدم چرا امروزاینقدر استرس و دلشوره دارم |
| Neutral | Neutral | Neutral | Neutral | I had asked the reason, but you hadn't answered, and now I understand the reason.<br>دلیلشو پرسیده بودم جواب نداده بودی که الان متوجه دلیلش شدم |
| Neutral | Positive | Neutral | Neutral | Now, for God's sake, I passed some of my wishes<br>حالا بخاطر خدا از بعضی خواسته هام گذشتم |

Dataset has been annotated by Five Persian native taggers. Each tagger chooses one of the three available tags as the best tag for each tweet. Some tweets may have complicated sentiments, and it can be challenging for a tagger to assign the best tag. In order to ensure a reliable sentiment tagging approach, we conducted several meetings with taggers throughout the annotation process to identify and rectify any mistake, thereby ensuring definition consistency between taggers and us. Some tweets are hard to assign a particular label and mostly do not have the same tag in annotation by different taggers. So these tweets were dropped because of the Unintelligibility of the text or disagreement between annotators. Finally, the tweets annotated by three taggers in the same class were selected as the dataset. Some samples of the proposed dataset have been exhibited in Table 2. Table 2 contains 5 columns, which include the text of the tweet with its translation into English, the 3 tags of the best taggers and the selected final tag assigned to the tweet respectively.

The Cohen's Kappa [24] measure has been calculated to evaluate the agreement between the annotators in assigning the tweets sentiment. Cohen's Kappa is a statistical measure used to assess the inter-rater reliability or agreement between two raters when categorizing data. Such calculations are important in various fields such as data annotation, where multiple annotators are involved. By analyzing the Kappa measures, we gain insights into the consistency and re liability of the annotations by different annotators. These calculations enhance our understanding of inter-rater reliability and show the quality and consistency of the annotations performed by the annotators. In order to have the accuracy of all annotators, Cohen's Kappa values were calculated for each tagger separately.

Table 3 displays the Kappa measures calculated for five annotator pairs. In this table each cell indicates the kappa calculated for two taggers(row and column names respectively) while the fifth Column shows the average Kappa value for five taggers(row-names). The column 5 provides an overview of the agreement between each pair of annotators. The obtained values show the agreement between the taggers so that the mean Kappa value for five taggers is equal to 66.1.

Table 3: The Cohen's Kappa agreement results for each tagger.

|  | Tagger1 | Tagger2 | Tagger3 | Tagger4 | Tagger5 |
|---|---|---|---|---|---|
| Tagger1 |  | 0.74 | 0.72 | 0.42 | 0.39 |
| Tagger2 |  |  | 0.76 | 0.74 | 0.74 |
| Tagger3 |  |  |  | 0.73 | 0.71 |
| Tagger4 |  |  |  |  | 0.66 |
| Mean | 0.66 | | | | |

To train and evaluate the models and compare the results of each model, the new dataset is split into train and test data. Splitting the data to train and test allows the researchers to compare the performance of different models on the same testing data and provide a fair comparison between them. The data is divided into train and test so that 11,000 tweets have been selected as training data and 1,000 tweets as test data. The data statistics details are summarized in Table 4.

Table 4: Dataset statistics details

| Classes | Train | Test |
|---|---|---|
| Positive | 2692 | 240 |
| Neutral | 4300 | 404 |
| Negative | 4008 | 357 |

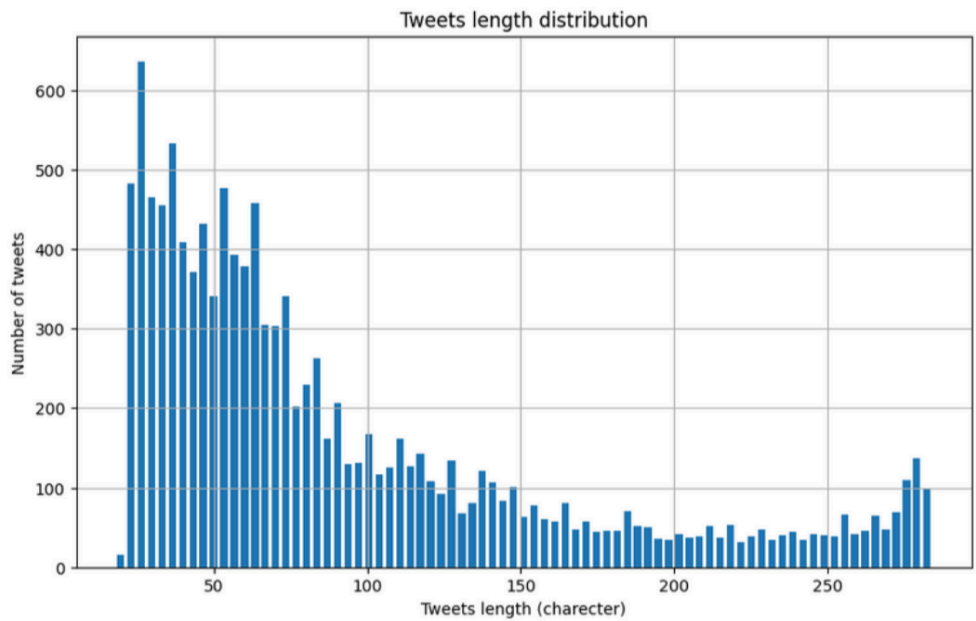

Figure 1: Tweets length distribution

Typically, users express their comments on Twitter through short texts. Figure 1 shows the distribution of tweets length before normalizing the training data. It shows that most of the tweets in the dataset are less than 150 characters long.

## 4 Preprocessing and Models

In this section, data preprocessing and models fine-tuning are discussed in detail. The significance of text preprocessing lies in its ability to minimize the amount of noise present in text data (this sentence needs reference). In text preprocessing, tweets are normalized in four steps described in sections 4.1. After that the normalized text is passed through a tokenization, vectorization and classification using a language model. In the classification step, two pre-trained language models ParsBERT [14] and RoBERTa [25], which were trained on Persian texts, are fine-tuned. More details about ParsBERT and RoBERTa and their fine-tuning with new sentiment dataset, are described in Section 4.2 and 4.3 respectively.

### 4.1 Data preprocessing

Tweets collected by the Twitter API generally contain texts with emails, links,

mentions, uncommon words, etc making them hard to classify. Another challenging case In Persian, is when users typically omit the use of half-space when expressing their opinions. In order to clean and normalize the tweets some steps are applied on texts. The following list outlines all pre-processing procedures that were applied on the dataset:

1. remove mention and links.

2. fix half-spaces, normalize tweet and numbers

3. remove non Persian characters.

4. fix punctuation

In these four steps, all links, mentions and non-Persian characters inside the tweet have been removed. All the cases where the user has mistakenly used a space, have been changed to a half-space (ex:" می توان" –> "می‌توان " which means "can do" in English). Afterwards, the numbers are normalized with the Hazm library (English numbers have been converted to Persian numbers). Finally, punctuation marks are separated from the words using a space.

### 4.2 Models

Language models trained on large amounts of data perform very well for tasks related to natural language understanding. Among these models, pre-trained transformer-based models such as BERT [26] and RoBERTa [25] have become increasingly popular due to their advanced performance. In this paper, Pars BERT [14] and Roberta, which are based on the Transformer architecture, are used on the collected Persian dataset.

**ParsBERT** BERT, a Bidirectional Transformer Encoder, has been trained on massive amounts of unlabelled text corpora using two training objectives: Masked Language Model (MLM) and Next Sentence Prediction (NSP). The MLM objective involves masking some words with probability of 15% within a sentence and predicting the masked words. In NSP, The training process involves using BERT to predict whether a given set of sentences are consecutive or not. Specifically, 50% of the sentences used for training are consecutive, while the other parts are random sentences. ParsBERT is based on BERT model architecture as a monolingual BERT for the Persian language. This model has been trained on various Persian collections such as news, novels, and scientific writings containing more than 2M documents. ParsBERT includes 100k initial Persian tokens, hidden and embedding layers with 768 dimensions, and 12 Transformer encoder layers with 12 attention heads in each layer. The ParsBert results indicate that it can achieve state-of-the-art performance compared to other architectures and multilingual models in Persian downstream tasks, such as Sentiment Analysis, Text Classification, and Named Entity Recognition.

**Roberta** Similar to BERT, RoBERTa (Robustly Optimized BERT Approach) is a transformer-based language model. Compared to BERT, Roberta trained the model longer on more data with bigger batches. The NSP objective is removed in RoBERTa, and MLM uses dynamic masking, where different parts of the sentences are masked in different epochs. This approach makes the model more robust.

For the Persian sentiment analysis task, Roberta must train on Persian data. For this purpose, Roberta, which contains six hidden layers, was used. This model has faster training and inference time than Roberta. The input length of tweets was set to a maximum of 256 tokens, because the text length of published Persian tweets is often in the form of short text. To train from scratch the Roberta, 50 million Persian tweets have been collected using the Twitter API. The pre-processing mentioned in sections 4.1 was applied to the collected tweets. The trained model includes 64k initial Persian tokens, a dimension of 768 for hidden and embedding layers, and 6 Transformer encoder layers with 6 attention heads.

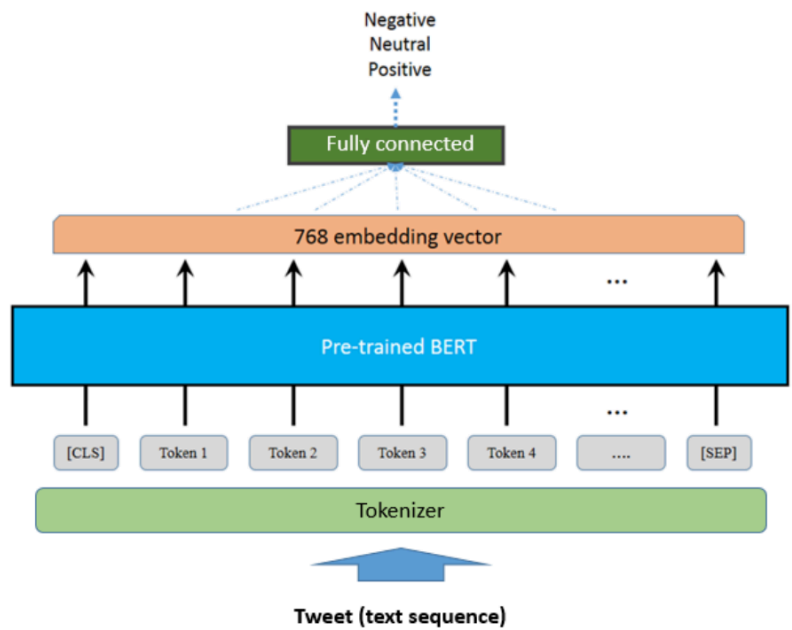

Figure 2: BERT architect with fully connected layer for sentiment analysis

### 4.3 Models Fine-tuning

Language models like BERT are pre-trained language models trained on massive amounts of text data. Fine-tuning BERT on a specific task involves taking the pre-trained model and training it further on a smaller, task-specific dataset. Fine-tuning allows the model to adapt its parameters and learn task-specific features, which can lead to improved performance on that particular task. By Fine-tuning,

BERT can achieve state-of-the-art results on these tasks with relatively small amounts of training data. A new sentiment analysis dataset is collected to fine-tune the models introduced in section 4.2. By fine-tuning the model on this dataset, It is expected that better results will be obtained on sentiment analysis, especially on Persian Twitter data.

Fine-tuning the models is done by adding a fully connected layer on top for classification. The input of a fully connected layer is a vector of 768 dimensions and its output is a vector of 3 dimensions that represents the output label of the sentence (as shown in Figure 2). As input data is fed, the entire pre-trained model and additional untrained classification layer are trained on sentiment analysis tasks.

One of the issues in ParsBERT fine-tuning is outdated its tokenizer. ParsBERT tokenizer does not have specific tokens for social networks like Twitter because it is not trained on social network data. Tokens such as emojis and important numbers (etc: "🥳 سال جدید ۲۰۴۱ مبارک " - Happy New Year 1402 🥳) are not available in this tokenizer. By adapting the tokenizer to the specific characteristics of Twitter data, the fine-tuned ParsBERT model can effectively capture the nuances of user-generated content, leading to more accurate and comprehensive language understanding. Therefore, for ParsBERT fine-tuning, all the emojis and important numbers have been added and updated ParsBERT tokenizer. After updating the tokenizer, ParsBERT with 12 block transformers and 12 attention heads is used for fine-tuning.

Roberta described in section 4.2 is used for fine-tuning with 6 block trans formers and 12 attention heads trained on Twitter data. Fine-tuning is done with batch size 32, max-length 256, Adam optimizer[27], and a learning rate of 5e-5 in 2 epochs. The dropout probability for all fully connected layers in the embeddings, encoder, and pooler is 0.1 for ParsBERT and Roberta. Fine-tuning the models is done on a 1080 GPU with 8G of RAM.

## 5 Discussion

The results of the models on test data are evaluated using three metrics called precision, recall and F1-score. These measures are mostly used to assess how well a classification is performed. The F1-score measures the harmonic mean of precision and recall which is more reliable than other metrics like accuracy especially when we have an imbalance dataset. This is because the accuracy of one class can get a high score and misclassify every other sample. Table 3 displays the results of the experiments on 1000 tweets as test data, where each row represents the precision, recall, and f1-score of a specific label for ParsBERT and Roberta. The mean f1-score of the three classes is 78.42 for ParsBERT and 79.87 for Roberta.

Table 5: Precision, Recall and F1-score of the models on new dataset

| Model | Label | Precision | Recall | F1-score |
|---|---|---|---|---|
| **ParsBERT** | Positive | 80.65 | 81.66 | 81.15 |
|  | Neutral | 75.05 | 78.21 | 76.60 |
|  | Negative | 79.82 | 75.35 | 77.52 |
|  | Macro avg | 78.51 | 78.41 | 78.42 |
| **Roberta** | Positive | 81.09 | 80.41 | 80.75 |
|  | Neutral | 78.85 | 78.46 | 78.66 |
|  | Negative | 79.77 | 80.67 | 80.22 |
|  | Macro avg | **79.90** | **79.85** | **79.87** |

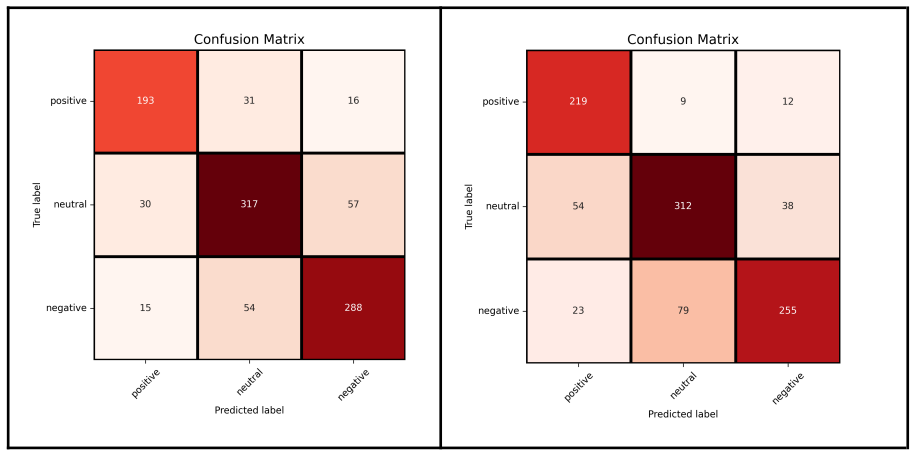

(a) Confusion matrix result for Roberta (b) Confusion matrix result for ParsBERT

Figure 3: Confusion matrix represented result on test data

3b) respectively. The results of Figure 3 and Table 5 show Roberta performs better than ParsBERT due to its pretraining on Twitter data from scratch and then fine-tuning for sentiment analysis. The result also highlights a slightly lower accuracy in detecting neutral tweets. It may be because of the presence of tweets for agency news and neutral idioms, which are neutral data from the point of view of data collectors and writer users, but the model cannot recognize them well.

# 6 Conclusion

This paper details the creation of a sentiment dataset in Persian. It encompasses the procedures involved in annotating Persian tweets with the assistance of annotators, and an exploration of some of the challenges encountered during the annotation process. Finally, it provides a summary of statistics related to Exa-PSD, including word counts in tweets, along with insights into inter annotator agreement and the results of fine-tuning a roberta language model on this data. Considering the characteristics of the Exa-PSD, this data can be useful for researchers interested in processing Persian contexts in the field of sentiment analysis. There are two important features in Exa-PSD making that unique compared to existing corpuses. First, this data contains about 12000 Per sian tweets that are annotated by 5 native taggers. This number is more than the size of other corpuses being annotated manually in Persian. In addition, our dataset is freely and publicly available for research activities as well. Future works could involve the incorporation of irony content when determining the sentiment labels of tweets, as numerous Twitter users employ irony extensively, particularly in Farsi content.

## Availability of Data and Materials

The datasets generated and analyzed during the current study are available from the corresponding author on reasonable request. Additionally, the dataset has been made publicly available at https://github.com/exaco/Exa-PSD, allowing for further research and validation of the findings.